\pdfoutput=1

\documentclass[11pt]{article}

\usepackage[final]{acl} 
\usepackage{times}
\usepackage{latexsym}
\usepackage[T1]{fontenc}
\usepackage[utf8]{inputenc}
\usepackage{microtype}
\usepackage{inconsolata}
\usepackage{graphicx}
\usepackage{lscape}
\usepackage{multirow}
\usepackage{listings}
\usepackage{pdfpages}
\usepackage{float}
\usepackage{tikz}
\usepackage{booktabs}
\usepackage{amsmath}

\title{Interaction Matters: An Evaluation Framework for Interactive Dialogue Assessment on English Second Language Conversations}

\author{Rena Gao \and Carsten Roever \and Jey Han Lau \\
University of Melbourne, Australia \\  
\texttt{wegao@student.unimelb.edu.au, \{carsten, laujh\}@unimelb.edu.au}
}

\begin{document}
\maketitle

\begin{abstract}
We present an evaluation framework for interactive dialogue assessment in the context of English as a Second Language (ESL) speakers. Our framework collects dialogue-level interactivity labels (e.g., topic management; 4 labels in total) and micro-level span features (e.g., backchannels; 17 features in total). Given our annotated data, we study how the micro-level features influence the (higher level) interactivity quality of ESL dialogues by constructing various machine learning-based models. Our results demonstrate that certain micro-level features strongly correlate with interactivity quality, like reference word (e.g., she, her, he), revealing new insights about the interaction between higher-level dialogue quality and lower-level fundamental linguistic signals. Our framework also provides a means to assess ESL communication, which is useful for language assessment\footnote{The dataset and code are available at: \url{https://github.com/RenaGao/2024InteractiveMetrics}}.
\end{abstract}

\section{Introduction}\label{sec:intro}
Estimates suggest more than 750 million individuals use English as a non-native language \cite{Dyvik_2023}. Despite its widespread use, a notable gap exists in the availability of datasets that capture the communicative features of English Second Language (ESL) speakers within dialogic contexts. Most existing dialogue datasets are primarily created with native speakers' conversations, failing to consider the distinct linguistic subtleties and obstacles encountered by ESL speakers \cite{settles2021epistemic} such as different usages on grammar, syntax and sentence structure influenced by their native languages. On the other hand, for dialogue quality evaluation, most existing performance metrics focus on fluency, coherence or consistency \cite{tao2018ruber}, which fail to capture or evaluate the sophisticated features of dialogue such as the speakers' ability to interact, manage topics through multi-turn dialogues, or use the appropriate tone given a particular domain/context. These gaps, in particular, are becoming more crucial due to the increasing demand to evaluate ESL speakers' communication and interaction skills, which is important not only for better cross-cultural exchanges but also for improving educational assessments. While resources such as the International Corpus of Learner English \cite{rica2009status} offer data from controlled spoken settings on monologic speech, they fall short in addressing multi-party interactive dialogues. 


\begin{figure}[t]
   \centering 
   \includegraphics[width=\linewidth]{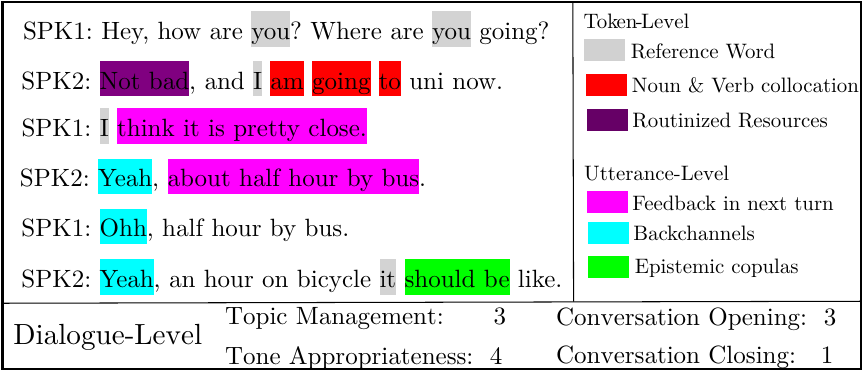}
   \caption{An example of an annotated dialogue with dialogue-level interactivity labels and micro-level features}\label{fig:inter-nature-conv-diag}
\end{figure}

In this paper, we introduce an ESL dialogue dataset and propose an evaluation framework designed to capture dialogue interactivity. Specifically, our framework has two different levels of annotation: (1) 4 dialogue-level interactivity labels that capture topic management, tone appropriateness and conversation opening and closing; and (2) 17 micro-level linguistic features that capture token-level features (e.g., reference word and routinized resources) and utterance-level features (e.g., epistemic copulas and backchannels). Figure~\ref{fig:inter-nature-conv-diag} illustrates an example of an annotated dialogue.
Appendix~\ref{subapp:MN} gives the full list of interactivity labels and micro-level features, along with their descriptions. Note that the micro-level features are annotated as spans, while the dialogue interactivity labels are document labels.

After annotating the ESL dialogues with our framework, we investigate the relationship between interactivity labels and micro-level features. To this end, we build machine learning models that use micro-level features as input to predict the interactivity labels of a dialogue.  We demonstrate how micro-level features impact various interactive aspects of ESL dialogues: specifically we saw which micro-level features contribute to the prediction of a particular interactivity quality.
Additionally, we also compare against a baseline BERT \cite{devlin2018bert} that uses the \textit{raw dialogue} as input to predict the interactivity labels, and found that it performs worse than our (simpler) machine learning models that use micro-level features as input, suggesting that these micro-level features have a stronger predictive power for interactivity. To summarize, our contributions are given as follows:
\begin{itemize}
\item We propose a novel evaluation framework for ESL dialogues that assesses four dialogue-level interactivity labels, including topic management, tone appropriateness and conversation opening and closing. It also captures seventeen fundamental micro-level features, such as backchannels (at the utterance-level) and reference words (at the token-level). 
\item We release SLEDE (\textbf{S}econd \textbf{L}anguage \textbf{E}nglish \textbf{D}ialogue \textbf{E}valuation), an annotated ESL dialogue dataset based on our evaluation framework. 
\item We study the interplay between the interactivity labels and micro-level features via predictive learning. Our experimental results explain how certain micro-level features impact various interactive aspects of ESL dialogues.
Our predictive models have the potential to be applied to real-world English tests to assess ESL communication.

\end{itemize}

\section{Related work}\label{sec:related-work}
\subsection{ESL Conversational Dialogue}

The interactive feature of human dialogue influences how turns and overlaps occur when analyzing conversations in communication, which is important for tagging and processing dialogue data \cite{allwood2008dimensions}.
Due to the complex nature of data collection and practical issues, open-source conversational dialogues are still limited in related research fields, and most conversational datasets are designed for speech recognition purposes \cite{lovenia2022clozer}. The interactive feature of conversations vary between English native speakers and ESL speakers. For native speakers, the fluidity and nuance of the language come naturally, allowing for a dynamic range of expressions and a deeper level of engagement in conversation. However, ESL speakers often navigate different social and cultural norms through the usage of a second language, which adds complexity and richness to the conversation dataset and reflects the multifaceted nature of human communication. Moreover, the learners' native languages frequently shape their learning and usage of a second language, resulting in distinct constructions, mistakes, and use patterns \cite{betts2003easyenglish, warren2017cross}. As a consequence, it is interesting to ask the following questions when creating a second language conversation dataset: (1) how can we annotate lower level grammar related and communicative features?; and (2) how can we capture the higher level dialogue interactivity qualities?

\subsection{Dialogue Interactivity Quality} \label{subsec:interactive-metrics}

Our evaluation framework assesses on four interactivity quality in dialogue: topic management, tone appropriateness, and conversation opening and closing. Here we discuss various studies focusing on these aspects, providing motivation on why we choose them in our framework.

\paragraph{Topic Management} How speakers collaboratively manage topics in a dialogue is an important indicator of interactional ability. Speakers exhibit increasing mutuality and engagement in their interactions \cite{galaczi2014interactional}. They demonstrate mutuality by taking up and extending interlocutor-initiated topics through reformulating interlocutor contributions \cite{lam2018counts}, and they provide frequent listener responses and assessments of interlocutor statements (``that's so cool'', ``definitely'', ``oh no''), thus creating a stronger sense of engagement \cite{galaczi2014interactional}. \citet{ghazarian2022deam} argued that evaluating topic coherence in human conversation is still a challenging task and called for a more empirical way of conducting this evaluation. 

\paragraph{Tone Appropriateness} \citet{10.1145/3424153} suggested the social role of a chatbot needs to be emphasised when measuring chatbot performance. As such, another important aspect of interactional ability is language choice following the social role. \citet{pill2016drawing} demonstrated the need for healthcare professionals to speak at a high level of linguistic proficiency and speak in ways particular to their profession. \citet{dai2022design} and \citet{dai2023promise} extended this work to other social roles and showed that language users are  capable of configuring their linguistic abilities to display attributes commonly associated with a particular social role in their interactions. \citet{roever2021reconceptualizing} and \citet{roever2023relationship} similarly found that humans learn to talk in ways conventionally expected for a social role. 
\paragraph{Conversation Opening and Closing} Opening and closing of conversations is a long-standing fundamental research concern in dialogues \cite{schegloff1968sequencing, schegloff1973opening}, which can also be used to differentiate levels of interactional ability. Proficient ESL speakers are found more likely to open the conversation with preliminary and affiliative talk than less proficient speakers \cite{abe2019interactional}. Similarly, proficient ESL speakers are shown to display more elaborate closings \cite{abe2020task}. \citet{stolcke2000dialogue}, however, argued that there is still a lack of practical measures to assess the performance of starting and closing a conversation.

\subsection{Dialogue Fundamental Features} 

There is a growing interest in developing more sophisticated evaluation frameworks that can adapt to the diverse grammatical structure of spoken interaction in dialogues \cite{sinha2020learning}, which is essential for understanding ESL communication. 
\citet{dinan2020second} argued that more advanced semantic analysis tools are needed to better understand vocabulary choices' impact on dialogue quality from a micro-level, including code-switching, response tokens, and tense choice for verbs. Currently, only limited works have discussed the empercial methods on how to link these vocabulary choices to demonstrate the quality of communication in conversations. 

For a bigger unit, utterance level features such as feedback in next turn and backchannels are all critical features in considering the quality of interactions \cite{wu2021proficiency}. In addition, notion in grammatical resources, such as modal verbs \cite{shaxobiddin2024discourse}, epistemic copulas \cite{hayashi2020gaze}, and collaborative finishes \cite{yap2024versatile}, highlight the ability in deploying basic fundamental resources in actual interaction when constructing a dialogue. 
Thus, the evaluation metrics need to be sensitive to the linguistic features of multiple languages and the contexts in which these choices occur. The 17 micro-level features in our framework are inspired by these studies.
\section{Evaluation Framework}\label{sec:span-annotation-framework}
\begin{figure}[t]
    \centering
    \includegraphics[width=\linewidth]{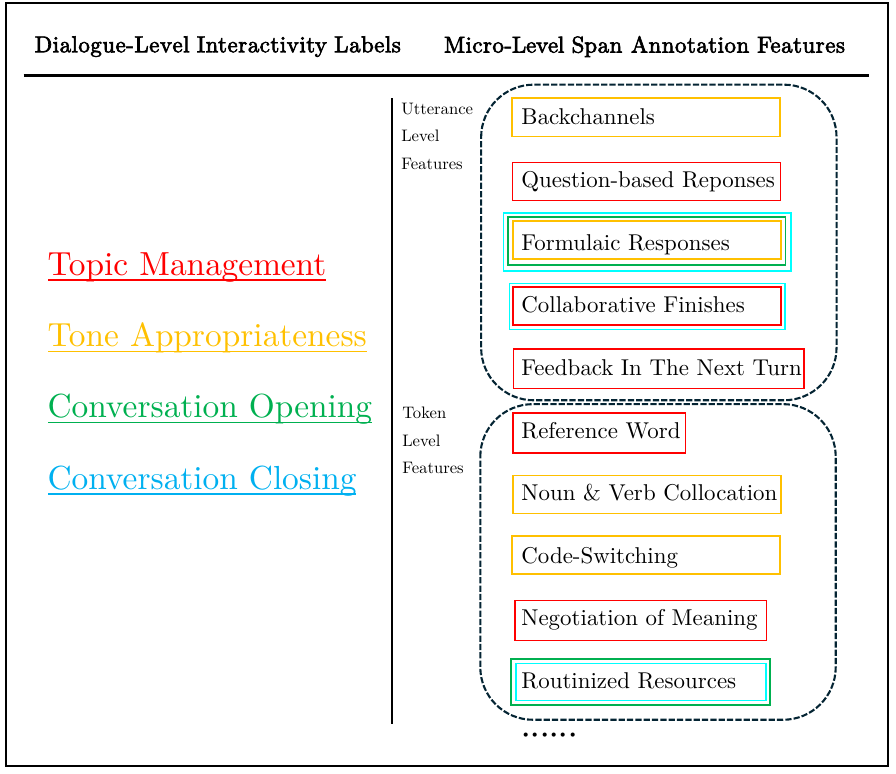}
    \caption{Our proposed evaluation framework has dialogue-level interactivity labels and micro-level features targeting interaction and engagement.}
    \label{fig:span-framework}
\end{figure}


Our motivation is to design a more comprehensive and transparent dialogue evaluation framework that captures dialogue interactivity and fundamental linguistics properties. To this end, we introduce an evaluation framework that has two levels of annotations: (1) dialogue-level interactivity labels (4 labels); and (2) micro-level linguistic features (17 features). 

\begin{table*}[ht]
\centering
\resizebox{1.0\textwidth}{!}{%
\begin{tabular}{lcl}
\toprule
\textbf{Interactivity Labels} & \textbf{Scores} &\textbf{Description of Scores} \\ 
\midrule
Topic Management &
\begin{tabular}[c]{@{}l@{}} 
  {[}5{]}\\  
  {[}4{]}\\
  {[}3{]}\\  
  {[}2{]}\\  
  {[}1{]}\\  
\end{tabular} &
\begin{tabular}[c]{@{}l@{}}
  topic extension with clear new context\\ 
  topic extension under the previous direction\\
  topic extension with the same content\\ 
  repeat and no topic extension\\ 
  no topic extension and stop the topic at this point\end{tabular}  \\ 
\midrule
Tone Appropriateness &
  \begin{tabular}[c]{@{}l@{}}
  {[}5{]}\\
  {[}4{]}\\
  {[}3{]}\\ 
  {[}2{]}\\ 
  {[}1{]}\\ 
  \end{tabular} &
  \begin{tabular}[c]{@{}l@{}}
  very informal \\ 
  quite informal, but some expressions are still formal \\ 
  relatively not formal, and most expressions are quite informal\\
  quite formal, and some expressions are not that formal\\
  very formal\end{tabular} \\ 
\midrule
Conversation Opening &
  \begin{tabular}[c]{@{}l@{}}
  {[}5{]}\\ 
  {[}4{]}\\ 
  {[}3{]}\\ 
  {[}2{]}\\ 
  {[}1{]}\\ 
\end{tabular} &
\begin{tabular}[c]{@{}l@{}}
nice greeting and showing a good understanding of the opening of conversation in social interactions. \\ 
sounded greeting and showed a basic understanding of the social role. \\
general greeting but not understanding the social role well. \\ 
basic greeting.  \\ 
no opening, start the discussion immediately.
\end{tabular} \\ 
\midrule
Conversation Closing &
\begin{tabular}[c]{@{}l@{}}
{[}5{]}\\ 
{[}4{]}\\ 
{[}3{]}\\ 
{[}2{]}\\ 
{[}1{]}\\ 
\end{tabular} &
\begin{tabular}[c]{@{}l@{}} detailed summarization and smooth transition to the closing of the conversation. \\ 
transit to the closing naturally, but without summarising the discussion. \\
transit to the discussion. \\ 
demonstrate a translation to the end of the conversation. \\ 
no closing, directly stop the conversation.
\end{tabular} \\ 
\bottomrule
\end{tabular}
}
\caption{Description of scores for dialogue-level interactivity labels. Higher score indicates better interactivity ability, for example, \textit{Tone Appropriateness} scores higher with more informality shows that the speakers are able to employ more active linguistics resources in dialogue communication to perform more causal and natural interactions compared with the formal tone, which has limited linguistics resources and not naturally occurred in real-life conversations.}
\label{tab:dialogue-feature-label}
\end{table*}

For the interactivity labels, we annotate: (1) {topic management}, which measures how extensively the topic is expanded upon and whether the content is new or previously discussed; (2) {tone appropriateness}, which indicates the degree of formality; (3) {conversation opening}, which rates the quality of greetings and (4) {conversation closing}, which rates the quality of summaries. Each of these labels is annotated with five categorical scores from 1 to 5 to assess the degree of interactivity; Table~\ref{tab:dialogue-feature-label} provides a detailed description for each score.

For micro-level features, we target grammatical, interactional and semantic aspects, and further decompose them into 7 token-level features that represent word formations that are indicative of an ESL speaker's ability to navigate linguistic resources for clarity, emphasis, and cultural relevance, including ``reference word'', ``noun \& verb collocation in proper form'', ``code-switching for communicative purposes'', ``negotiation of meaning'', ``tense choice to indicate interactive aims'', ``routinized resources'' and ``subordinate clauses'' ; and 10 utterance-level features for contextual interactions including ``backchannels''; ``responses framed as questions''; ``formulaic expressions''; ``collaborative finishes''; ``adjectives and adverbs denoting possibility''; ``constructions with impersonal subjects'' followed by ``non-factive verbs and noun phrases'' and ``feedback in the next turn'', ``impersonal subject + non-factive verb + NP'', ``adjectives/ adverbs expressing possibility''. These features capture the dynamic interplay between speakers, emphasizing the importance of backchannels, question-framed responses, and other mechanisms that facilitate a collaborative and adaptive exchange. Figure~\ref{fig:span-framework} summarises our evaluation framework, and 
Appendix~\ref{subapp:MN} provides the full details of these labels/features.

\section{SLEDE Development}\label{sec:dataset}
We now describe how we develop SLEDE (Second Language English Dialogue Evaluation): i.e.\ how we collect ESL dialogue data (Section~\ref{subsec:ESL-collect}) and annotate the data based on our evaluation framework (Section~\ref{subsec:annotate}).

\subsection{ESL Dialogue Collection} \label{subsec:ESL-collect}

We first look at finding the right set of conversational topics for the participants. We came up with a preliminary set of topics, and survey a group of 60 individuals, comprising both native English speakers and ESL speakers, to get their feedback on the quality of the topics. After collecting the feedback, we used their insights to further refine the topic set; the final set of topics are presented as part of the questionnaire that we ask participants to fill in before we collect their dialogues (``block 4'' in Appendix \ref{subapp:QA}). 

Next, we recruit 120 Chinese ESL speakers (volunteers) to engage in a 1-to-1 in-person talk on a chosen topic. The criteria for selecting volunteers for collecting the datasets are given as follows: (1) An IELTS score exceeding 6.5 to comprehend the dialogue fully; (2) A minimum educational attainment of a bachelor's degree in data science, computer science, or linguistics from a recognized university;
(3) Consent to agree on recording (refer to Appendix~\ref{subapp:SI} for details). These prerequisites were established to guarantee that the workers possess proficient English comprehension and are adequately equipped to have a high-quality conversation for the pair discussion. All speakers will then go through a training phase to ensure they understand the task. We follow \citet{mehri2022interactive} where we provide instructions (Appendix~\ref{subapp:SI}) to highlight important dialogue aspects to take into account, such as coherence, language complexity, and naturalness.

After training, we break the 120 volunteers into 60 pairs. The pair was matched with similar second language proficiency to ensure the dialogue maintains a stable quality within the two speaker's interaction, for example, a IELTS 6 ESL speaker was paired with another IELTS 6.5 speaker. Each pair undergoes two rounds of conversation: the first half-hour is dedicated to discussing a specific topic (chosen by them in the questionnaire), and the second half-hour involves discussing a specific issue and proposing solutions (Appendix~\ref{subapp:SI}).
We therefore collected a total of 120 dialogues, each lasting about half an hour, with thousands of turns in each dialogue.

\subsection{Dialogue-level Interactivity Label and Micro-level Feature Annotation} 
\label{subsec:annotate}

Given the 120 dialogues, we now collect annotations based on our proposed evaluation framework (Section~\ref{sec:span-annotation-framework}). To this end, we recruit eleven volunteer postgraduate students proficient in English (six in computer science and four in applied linguistics). These eleven annotators and the first author were randomly split into six pairs to annotate the dialogues.

\begin{table}[t]
\centering
\resizebox{\columnwidth}{!}{
\begin{tabular}{lll}
\hline
Datasets &  \begin{tabular}[c]{@{}l@{}}Full\\ dialogues\end{tabular} \\ 
\hline
\# dialogues     & 120   \\
\# turns (max)   & 2,065 \\ 
\# turns (avg)    & 1,760 \\ 
\hline
\# words marked (token-level features)     & 10,852 \\
\# words marked (utterance-level features)  & 3,516  \\
\# micro-level features (total counts)      & 14,386   \\
\hline
\end{tabular}}
\caption{Annotated Data Statistics} \label{tab:dataset-statistics}
\end{table}

The annotators were presented with an annotation guide (Appendix~\ref{subapp:MN}) to explain the dialogue-level interactivity labels and micro-level features. It includes definitions and examples of each label/feature, as well as guidelines for using the annotation interface. For the dialogue-level interactivity labels (topic management, tone appropriateness, conversation opening and closing), the annotators are asked to give a score for each of the four labels, and the task is framed as a document labelling task. We adopt a majority voting approach to annotate labels, e.g., if annotators give different labels to the same dialogue, we select the most frequent label as our final label. For micro-level features, they are framed as a span annotation task where the annotators are asked to mark word spans that exhibit a particular micro-level feature. Note that a word can be marked with multiple features. 

To ensure the quality of the annotation process, annotators went through a training process where they were first asked to label six pilot dialogues, and the first author cross-checks all annotations. Any mistakes are then discussed. After the training, each pair of annotators (including the first author) are given 30 dialogues to annotate (noting that there is some overlapping dialogues between pairs). In total, 120 dialogues are annotated; some statistics of the annotated dataset are presented in Table ~\ref{tab:dataset-statistics}.

To understand annotation quality, we compute inter-annotator agreement for the interactivity labels and micro-level features. For the interactivity labels, we compute agreement between the annotators in a pair and take the average across the pairs.
For the micro-level features, we again measure agreement between the annotators in a pair at the token-level for each micro-level feature --- i.e., we first break the dialogue into individual word tokens and compute statistics based on the presence or absence of the feature as marked by the annotators for each word token\footnote{In other words, the unit of analysis here is a word token, and the output is a binary value for each annotator indicating whether it has been marked for the feature.} --- before aggregating over the features and pairs. We calculate Pearson correlation coefficient $r$ \cite{cohen2009pearson} and Krippendorff's $\alpha$ \cite{krippendorff2018content} to measure inter-annotator agreement, and the results are summarized in Table~\ref{tab:annotator-agreement}. The agreement is above 0.6 for micro-level features (token-level and utterance-level) and dialogue-level labels, indicating that there is a good consensus among annotators and the evaluation framework is robust/reliable. 

\begin{table}[t]
\centering
\resizebox{\columnwidth}{!}{
\begin{tabular}{cccc}
\toprule
\textbf{Measure} &
\textbf{Token-level} &  \textbf{Utterance-level} &  \textbf{Dialogue-level}  \\ 
&\textbf{Features} & \textbf{Features} & \textbf{Labels} \\
\midrule
 $\alpha$ & {0.63}  &{0.64}  & {0.65}   \\ 
 $r$ & {0.64}    & {0.67}   &{0.68}  \\
\bottomrule
\end{tabular}}
\caption{Inter-annotator agreement for micro-level features (token-level and utterance-level) and dialogue-level labels.} \label{tab:annotator-agreement}
\end{table}


\section{Experiments}\label{sec:experiment}
We conduct a series of experiments to analyse the influence of micro-level features on dialogue-level interactivity labels. To this end, we first build machine learning models to evaluate the prediction performance of interactivity labels given micro-level features as input in Section~\ref{subsec:model-analysis}, and then analyse the importance of micro-level features in Section~\ref{subsec:feature-importance} and lastly look at the difference between utterance-level vs.\ token-level features in Section~\ref{subsec:ablation}.

Given that our ESL dialogues are very long (maximum of 2065 turns as shown in Table \ref{tab:dataset-statistics}) and we only have a small number of them (120 dialogues),  we break each dialogue into smaller ``mini-dialogues'' that have a maximum of 12 turns in our experiments. This process produces 625 mini dialogues in total. For the micro-level labels, we can carry across the annotations we have collected for the original dialogues. For the interactivity labels, however, we copy the original labels from the larger dialogue they belong to.
To measure the validity of this approach, we randomly sample 60 mini-dialogues and re-annotate them (with 6 annotators) for the interactivity labels. Then, we measure the correlation between the two judgements (i.e., judgements copied from the original dialogues vs.\ judgements collected using mini-dialogues). We found the Pearson correlation to be 0.72, suggesting that our approach of copying the interactivity labels from the larger dialogue is a sensible way of creating labels for the mini-dialogues. Henceforth, all experiments that we describe use the mini-dialogues.

\subsection{Predicting Dialogue interactivity labels} \label{subsec:model-analysis}
We experiment with three machine learning algorithms, logistic regression (LR), random forest (RF), and Naïve Bayes (NB),
for predicting each dialogue interactivity label using the micro-level features as input. We frame this as a classification problem, where the model needs to output one of the five classes. For each micro-level feature, the feature weight ($x$) of a 
 mini-dialogue is computed as a weighted average of the fraction of marked tokens over the annotators:
\begin{equation}
    x = \sum_{i=1}^{N} \frac{c_i}{\sum_{j=1}^N c_j} \times \frac{c_i}{c_{\mathrm{total}}}
\end{equation}
where $N$ is the number of annotators who worked on the mini-dialogue, $c_i$ the number of marked word tokens by annotator $i$, and $c_{\mathrm{total}}$  the total number of word tokens in the mini-dialogue. Intuitively, we give more weights to annotators who highlight more words than those who highlight less, and the rationale for doing this is that \textit{under-marking} is a type of mistake more prevalent than \textit{over-marking}, based on a preliminary analysis of the data (and so annotators who don't mark many words should be down-weighted, as their annotations are likely to be of lower quality).

We also include a baseline, where we fine-tune BERT using the \textit{raw dialogue} as input to predict the interactivity labels.\footnote{We use `bert-base-uncased'.} This baseline tells us whether the micro-level features are actually useful, or we can use the raw dialogues directly for predicting the interactivity labels. We summarize our results in Table~\ref{tab:prediction-result} over four metrics: accuracy (ACC), precision (PRE), recall (REC), and F1 Score (F1).

\begin{table}[t]
\centering
\begin{tabular}{rcccc}
\hline
\multicolumn{5}{c}{Classification Models}\\ 
\hline
Labels & Topic & Tone  & Opening & Closing  \\ 
\hline
\multicolumn{5}{c}{\textbf{Logistic Regression}}\\
\hline
ACC & 0.815 & 0.849 & 0.975 & 0.950 \\
PRE & 0.690 & 0.746 & 0.950 & 0.941 \\
REC & 0.815 & 0.849 & 0.975 & 0.950 \\
F1 & 0.747 & 0.794 & 0.962 & 0.945 \\
\hline
\multicolumn{5}{c}{\textbf{Random Forest}}\\ 
\hline
ACC & 0.832 & 0.832 & 0.966 & 0.966 \\
PRE & 0.714 & 0.744 & 0.950 & 0.934 \\
REC & 0.832 & 0.832 & 0.966 & 0.966 \\
F1  & 0.766 & 0.786 & 0.958 & 0.950 \\
\hline
\multicolumn{5}{c}{\textbf{Naïve Bayes}}\\ 
\hline
ACC & 0.807 & 0.840 & 0.966 & 0.958 \\
PRE & 0.688 & 0.733 & 0.950 & 0.934 \\
REC & 0.807 & 0.840 & 0.966 & 0.958 \\
F1  & 0.743 & 0.783 & 0.958 & 0.946 \\
\hline
\multicolumn{5}{c}{\textbf{BERT}}\\ 
\hline
ACC & 0.528 & 0.530 & 0.719 & 0.746 \\
PRE & 0.519 & 0.609 & 0.647 & 0.682 \\
REC & 0.617 & 0.584 & 0.708 &  0.713 \\
F1  & 0.572 & 0.620 & 0.733 & 0.752 \\
\hline
\end{tabular}
\caption{The classification prediction results with different performance metrics accuracy (ACC), precision (PRE), recall (REC) and f1 score (F1) on the SLEDE dataset.} \label{tab:prediction-result}
\end{table}


\begin{table*}[t]
\centering
\begin{tabular}{ccc}
\toprule
\textbf{LR} &              \textbf{RF} &                      \textbf{NB} \\
\midrule
\textbf{Code Switching} & \textbf{Code Switching} & \textbf{Feedback in Next Turn*}\\ 
\textbf{Reference Word*} & \textbf{Feedback in Next Turn*} & \textbf{Formulaic Responses} \\
\textbf{Feedback in Next Turn*} & Question-based responses & \textbf{Reference Word*} \\
\textbf{Formulaic Responses}  & Non-factive Verb & Negotiation of Meaning \\ 
\textbf{Tense Choice} & \textbf{Reference Word*} & \textbf{Tense Choice} \\ 
\bottomrule
\end{tabular}
\caption{High impact common micro-level features over the three classifiers for predicting dialogue-level labels. Bold/asterisk indicates overlapping features in two/three classifiers.} \label{tab:common_features}
\end{table*}

\begin{table*}[h]
\centering
\resizebox{2\columnwidth}{!}{
\begin{tabular}{cccc}
\toprule
\textbf{Topic} & \textbf{Tone} & \textbf{Opening} & \textbf{Closing} \\ 
\midrule
\multicolumn{4}{c}{\textbf{Logistic Regression}} \\
\midrule
\textbf{Negotiation of Meaning*} & \textbf{Routinized Resources*} & Epistemic Modals & \textbf{Backchannels*}\\
\textbf{Subordinate Clauses*} & Adj./Adv. Expressing & \textbf{Formulaic Responses} & \textbf{Adj./Adv. Expressing} \\
Noun\&Verb Collocation & \textbf{Feedback in Next Turn*} & \textbf{Question-Based Responses*} & Formulaic Responses \\
\textbf{Question-Based Responses} & \textbf{Formulaic Responses*} & \textbf{Subordinate Clauses*} & \textbf{Collaborative Finishes*} \\
\textbf{Negotiation of Meaning} &\textbf{Reference Word}  & \textbf{Adj./Adv. Expressing*} & Epistemic Copulas \\
\midrule

\multicolumn{4}{c}{\textbf{Naïve Bayes}} \\
\midrule
non-factive verb phrase structure & \textbf{Routinized Resources*} & \textbf{Adj./Adv. Expressing*} & \textbf{Adj./Adv. Expressing} \\
\textbf{Question-Based Responses} & \textbf{Feedback in Next Turn*} & \textbf{Routinized Resources} & Epistemic Modals \\
Adj./Adv. Expressing & \textbf{Epistemic Copulas} & \textbf{Subordinate Clauses*} & \textbf{Backchannels*} \\
\textbf{Negotiation of Meaning*} & Question-Based Responses & \ Epistemic Copulas  & \textbf{Collaborative Finishes*}  \\
\textbf{Subordinate clauses*} & \textbf{Subordinate Clauses*} & \textbf{Question-Based Responses*} & Question-Based Responses \\
\midrule

\multicolumn{4}{c}{\textbf{Random Forest}} \\
\midrule
\textbf{Negotiation of Meaning*} & \textbf{Epistemic Copulas} & Feedback in Next Turn & Feedback in Next Turn \\
Formulaic Responses & {Backchannels} & \textbf{Subordinate Clauses*} &Subordinate clauses  \\
\textbf{Subordinate Clauses*} & \textbf{Feedback in Next Turn*} & \textbf{Adj./Adv. Expressing*} & \textbf{Collaborative Finishes*} \\
Epistemic Copulas & \textbf{Negotiation of Meaning} & \textbf{Question-Based Responses*} &  \textbf{Formulaic Responses} \\
\textbf{Question-Based Responses} & \textbf{Routinized Resources*} & \textbf{Formulaic Responses} & \textbf{Backchannels*} \\ \midrule
\end{tabular}
}
\caption{High impact interactivity-specific micro-level features. For each interactivity label, bold/asterisk indicates overlapping features in two/three classifiers.}\label{tab:your_label}
\end{table*}

From the results, we observe that the three simple models (LR, RF, and NB) perform exceptionally well on conversation opening and closing, often achieving or nearing 0.95 and above for all metrics, indicating that these labels are easier to predict as they only appear at the beginning and the end of the conversation. Topic management and tone prediction, on the other hand, has a lower performance, and it is unsurprising given that it is arguably a more difficult task. That said, we're still seeing over 75\% F1 performance in most cases, suggesting that the micro-level features predictive of these interactivity labels.

Interestingly, BERT consistently underperforms by a large margin compared to the simple models. This implies the micro-level features are more predictive of the interactivity labels, and the raw dialogue alone does not provide the same level of information and pretraining isn't good enough close the gap.


Looking at the differences between classifiers, we see largely similar/consistent results, suggesting that the predictive performance is agnostic to the exact implementation of the classifier. We want to note that due to the lack of other ESL conversation datasets, these classifiers are trained from scratch (without having any form of pretraining). Compared to previous studies that found poor performance in classifying topics \cite{stolcke2000dialogue} and tone choices \cite{ghazarian2022deam} our results are encouraging. 



Taking all these observations together, given the relatively strong classification performance, the main insight we can draw here is that the micro-level features are able to explain the four dialogue interactivity qualities, shedding light into the possibility of using this interactive framework in the evaluation of dialogue beyond the ESL context. That is, one future direction for developing dialogue evaluation metrics is to consider incorporating some of these micro-level token and utterance features.


\subsection{Feature Importance Analysis} \label{subsec:feature-importance}
In this section, we further examined the significance of token and utterance-level features for predicting dialogue interactivity, aiming to identify the most important linguistic features influencing different interactive perspectives. This approach may provide insights into the foundational elements that drive the dialogue engagement.

Given that a trained LR, NB and RF classifier all provide weights to indicate the importance of each feature, for each classifier we first compute \textit{common} micro-level features $f_{\mathrm{c}}$ across the four interactivity labels:
\begin{align*}
f_{\mathrm{c}} = &\mathrm{top5}\big(\mathrm{top10}(f_{\mathrm{topic}}) \cap \mathrm{top10}(f_{\mathrm{tone}})  \\
&\cap \mathrm{top10}(f_{\mathrm{opening}}) \cap \mathrm{top10}(f_{\mathrm{closing}})\big)
\end{align*}
where $\mathrm{topk}$ is a function that returns the best $k$ items given by their weights, $f_{\mathrm{topic}}$ denote the set of micro-level features with their weights for predicting the topic management interactivity label. 

We display these common features in Table~\ref{tab:common_features} for LR, RF and NB. Interestingly, for common Top-5 features across three models,
we observed that ``Code-Switching'', ``Reference Word'', and  ``Tense Choice'' are shared across all classifiers (asterisk), and ``Formulaic Responses'', ``Code Switching'', ``Feedback in Next Turn'' is common across two out of three classifiers. We see very consistent highly impact common micro-level features over these different classifiers, suggesting that these features are reliable for predicting the interactivity labels. From the perspective of linguistic constructs for interactive purpose, ``Reference Word'' indicates the proper person to refer to in dialogue construction \cite{roever2023relationship}; for ``Code-Switching'', and  ``Tense Choice'' demonstrate the ability of smoothing the communication for second language speakers. ``Feedback in Next Turn'' presents the awareness of giving immediate responses in-time, which is essential in ensuring the dialogue quality in second language interaction. 

We next look at micro-level features that are specific to each of the interactivity label. To that end, for each classifier we compute interactivity-specific features, e.g., for topic management, as follows:
\begin{equation}
    \mathrm{top10}(f_{\mathrm{topic}}) - f_c
\end{equation}

Results for presented in Table~\ref{tab:your_label}. Again, we see consistent results between classifiers for each interactivity label. Many of these interactivity-specific micro-level features are intuitive. For example, for topic management, we have ``Negotiation of Meaning'' and ``Subordinate Clauses'' because these micro-level features tell us about the content and discourse of the discussion and indicate the transitions for topics. For tone appropriateness, ``Routinized Resources'', ``Formulaic Responses'', and ``Feedback in Next Turn'' are essential to demonstrate the social role in language resources choice. For conversation opening, ``Question-Based Responses'',``Subordinate Clauses'', and ``Adj./Adv. Expressing'' all show how a dialogue will be started from both speakers. And lastly for conversation closing, ``Collaborative Finishes'' and ``Formulaic Responses'' are directly link to the development of how to end a dialogue. 

To conclude, the largely consistent results between classifiers suggest that our findings are robust and not sensitive to the implementation of the classifier. That said, the fact there is some (minor) difference does suggest that there is perhaps complementarity between these classifiers and points to a potential future direction of ensembling these classifiers to improve the prediction of interactivity labels.

\subsection{Ablation Study}\label{subsec:ablation}
We now examine the individual effects of utterance-level and token-level features in the learning model predictions for the four interactivity qualities. As before, we train three classifiers (LR, RF, and NB) but this time they use only either the token-level (``Token'') or utterance-level (``Utt.'') features; results are presented in Table~\ref{tab:prediction-result-ablation}. Note that we also include the original results using both sets of features (``Both'') for comparison. The results indicate that predictions at the token level are better than those at the utterance level, though the difference isn't large.
%
%
Perhaps most importantly, we see that using both features together produce the best performance in most of the cases (exceptions: RF for Topic and Tone), showing that both types of micro-level features are important for predicting the dialogue-level interactivity labels.

\begin{table}[t]
\centering
\resizebox{\columnwidth}{!}{
\begin{tabular}{cc@{\;\;}c@{\;\;}cc@{\;\;}c@{\;\;}c}
\toprule
\textbf{Models} & \textbf{Token} & \textbf{Utt.}  & \textbf{Both}  &  \textbf{Token} & \textbf{Utt.} & \textbf{Both}  \\ 
\midrule
 & \multicolumn{3}{c}{Topic} &  \multicolumn{3}{c}{Tone}\\
\midrule 
LR    & 0.571 & 0.658 & 0.747 & 0.690 & 0.609 &  0.794  \\
RF    & 0.888 & 0.799 & 0.766 & 0.911 & 0.898  & 0.786 \\
NB    & 0.589 & 0.576 & 0.743 & 0.680 & 0.673  & 0.783  \\
\midrule
 & \multicolumn{3}{c}{Opening} &  \multicolumn{3}{c}{Closing}\\
\hline 
LR  & 0.915 & 0.840  & 0.962 & 0.934 &  0.711 & 0.945 \\
RF  & 0.974 & 0.978  & 0.958 & 0.976 &  0.981 & 0.950 \\
NB  & 0.915 & 0.914  & 0.958 & 0.934 &  0.928 & 0.946 \\
\bottomrule
\end{tabular}
}
\caption{The F1 results with different machine learning models across different feature levels.} \label{tab:prediction-result-ablation}
\end{table}

\section{Conclusions}\label{sec:conclusion}
In this paper, we propose a novel evaluation framework to assess the dialogue quality of ESL conversations by considering high-level interactivity labels and micro-level linguistic features. We develop SLEDE, an annotated ESL dialogue corpus based on the evaluation framework. We found that the micro-level features are highly predictive of the interactivity labels, and revealed impactful micro-level features that are: (1) common across different interactivity labels; and (2) specific to a particular interactivity label. Our results provide new insights into educational assessment for ESL communication.


\section{Limitations}
The developed dataset is admittedly small (120 dialogues). That said, the quality of the annotation is high (strong annotator agreement) and each dialogue is very long (almost 1,800 turns on average per dialogue). Ultimately, as our goal is to analyse the relationship between micro-level vs.\ interactivity features, our predictive models do not offer an end-to-end approach for evaluating dialogue quality, as it requires micro-level features as input. The future work will extend the scope by automating the processing for micro-level features. 

\section*{Ethics Statement}
This study is conducted under the guidance of the ACL Code of Ethics. We manually filtered out potentially offensive content and removed all information related to the identification of annotators. The annotation protocol is
approved under the The University of Melbourne's Human Ethics Application (reference number The University of Melbourne Human Ethics LNR as 2022-24988-32929-3.).



\bibliography{custom}

\begin{thebibliography}{32}
\expandafter\ifx\csname natexlab\endcsname\relax\def\natexlab#1{#1}\fi

\bibitem[{Abe and Roever(2019)}]{abe2019interactional}
Makoto Abe and Carsten Roever. 2019.
\newblock Interactional competence in l2 text-chat interactions: First-idea
  proffering in task openings.
\newblock \emph{Journal of Pragmatics}, 144:1--14.

\bibitem[{Abe and Roever(2020)}]{abe2020task}
Makoto Abe and Carsten Roever. 2020.
\newblock Task closings in l2 text-chat interactions: A study of l2
  interactional competence.
\newblock \emph{Calico Journal}, 37(1):23--45.

\bibitem[{Allwood(2008)}]{allwood2008dimensions}
Jens Allwood. 2008.
\newblock Dimensions of embodied communication-towards a typology of embodied
  communication.
\newblock \emph{Embodied communication in humans and machines}, pages 257--284.

\bibitem[{Betts(2003)}]{betts2003easyenglish}
Robert Betts. 2003.
\newblock Easyenglish: Challenges in cross-cultural communication.
\newblock In \emph{EAMT Workshop: Improving MT through other language
  technology tools: resources and tools for building MT}.

\bibitem[{Cohen et~al.(2009)Cohen, Huang, Chen, Benesty, Benesty, Chen, Huang,
  and Cohen}]{cohen2009pearson}
Israel Cohen, Yiteng Huang, Jingdong Chen, Jacob Benesty, Jacob Benesty,
  Jingdong Chen, Yiteng Huang, and Israel Cohen. 2009.
\newblock Pearson correlation coefficient.
\newblock \emph{Noise reduction in speech processing}, pages 1--4.

\bibitem[{Dai(2022)}]{dai2022design}
David~Wei Dai. 2022.
\newblock \emph{Design and validation of an L2-Chinese interactional competence
  test}.
\newblock Ph.D. thesis, University of Melbourne (Australia).

\bibitem[{Dai and Davey(2023)}]{dai2023promise}
David~Wei Dai and Michael Davey. 2023.
\newblock On the promise of using membership categorization analysis to
  investigate interactional competence.
\newblock \emph{Applied Linguistics}.

\bibitem[{Devlin et~al.(2018)Devlin, Chang, Lee, and
  Toutanova}]{devlin2018bert}
Jacob Devlin, Ming-Wei Chang, Kenton Lee, and Kristina Toutanova. 2018.
\newblock Bert: Pre-training of deep bidirectional transformers for language
  understanding.
\newblock \emph{arXiv preprint arXiv:1810.04805}.

\bibitem[{Dinan et~al.(2020)Dinan, Logacheva, Malykh, Miller, Shuster, Urbanek,
  Kiela, Szlam, Serban, Lowe et~al.}]{dinan2020second}
Emily Dinan, Varvara Logacheva, Valentin Malykh, Alexander Miller, Kurt
  Shuster, Jack Urbanek, Douwe Kiela, Arthur Szlam, Iulian Serban, Ryan Lowe,
  et~al. 2020.
\newblock The second conversational intelligence challenge (convai2).
\newblock In \emph{The NeurIPS'18 Competition: From Machine Learning to
  Intelligent Conversations}, pages 187--208. Springer.

\bibitem[{Dyvik(2023)}]{Dyvik_2023}
Einar~H. Dyvik. 2023.
\newblock \href
  {https://www.statista.com/statistics/266808/the-most-spoken-languages-worldwide/}
  {The most spoken languages worldwide 2023}.

\bibitem[{Galaczi(2014)}]{galaczi2014interactional}
Evelina~D Galaczi. 2014.
\newblock Interactional competence across proficiency levels: How do learners
  manage interaction in paired speaking tests?
\newblock \emph{Applied linguistics}, 35(5):553--574.

\bibitem[{Ghazarian et~al.(2022)Ghazarian, Wen, Galstyan, and
  Peng}]{ghazarian2022deam}
Sarik Ghazarian, Nuan Wen, Aram Galstyan, and Nanyun Peng. 2022.
\newblock Deam: Dialogue coherence evaluation using amr-based semantic
  manipulations.
\newblock \emph{arXiv preprint arXiv:2203.09711}.

\bibitem[{Hayashi(2020)}]{hayashi2020gaze}
Yugo Hayashi. 2020.
\newblock Gaze awareness and metacognitive suggestions by a pedagogical
  conversational agent: an experimental investigation on interventions to
  support collaborative learning process and performance.
\newblock \emph{International Journal of Computer-Supported Collaborative
  Learning}, 15(4):469--498.

\bibitem[{Krippendorff(2018)}]{krippendorff2018content}
Klaus Krippendorff. 2018.
\newblock \emph{Content analysis: An introduction to its methodology}.
\newblock Sage publications.

\bibitem[{Lam(2018)}]{lam2018counts}
Daniel~MK Lam. 2018.
\newblock What counts as “responding”? contingency on previous speaker
  contribution as a feature of interactional competence.
\newblock \emph{Language Testing}, 35(3):377--401.

\bibitem[{Lovenia et~al.(2022)Lovenia, Wilie, Chung, Zeng, Cahyawijaya, Dan,
  and Fung}]{lovenia2022clozer}
Holy Lovenia, Bryan Wilie, Willy Chung, Min Zeng, Samuel Cahyawijaya, Su~Dan,
  and Pascale Fung. 2022.
\newblock Clozer: Adaptable data augmentation for cloze-style reading
  comprehension.
\newblock \emph{arXiv preprint arXiv:2203.16027}.

\bibitem[{Mehri et~al.(2022)Mehri, Feng, Gordon, Alavi, Traum, and
  Eskenazi}]{mehri2022interactive}
Shikib Mehri, Yulan Feng, Carla Gordon, Seyed~Hossein Alavi, David Traum, and
  Maxine Eskenazi. 2022.
\newblock Interactive evaluation of dialog track at dstc9.
\newblock \emph{arXiv preprint arXiv:2207.14403}.

\bibitem[{Pill(2016)}]{pill2016drawing}
John Pill. 2016.
\newblock Drawing on indigenous criteria for more authentic assessment in a
  specific-purpose language test: Health professionals interacting with
  patients.
\newblock \emph{Language Testing}, 33(2):175--193.

\bibitem[{Rica-Peromingo(2009)}]{rica2009status}
Juan-Pedro Rica-Peromingo. 2009.
\newblock \emph{The Status of English in Spain}, pages 168--174.

\bibitem[{Roever and Dai(2021)}]{roever2021reconceptualizing}
Carsten Roever and David~W Dai. 2021.
\newblock Reconceptualizing interactional competence for language testing.
\newblock \emph{Assessing speaking in context: Expanding the construct and its
  applications}, pages 23--49.

\bibitem[{Roever and Ikeda(2023)}]{roever2023relationship}
Carsten Roever and Naoki Ikeda. 2023.
\newblock The relationship between l2 interactional competence and proficiency.
\newblock \emph{Applied Linguistics}, page amad053.

\bibitem[{Schegloff(1968)}]{schegloff1968sequencing}
Emanuel~A Schegloff. 1968.
\newblock Sequencing in conversational openings 1.
\newblock \emph{American anthropologist}, 70(6):1075--1095.

\bibitem[{Schegloff and Sacks(1973)}]{schegloff1973opening}
Emanuel~A Schegloff and Harvey Sacks. 1973.
\newblock Opening up closings.

\bibitem[{Settles et~al.(2021)Settles, Jones, Buchanan, and
  Dotson}]{settles2021epistemic}
Isis~H Settles, Martinque~K Jones, NiCole~T Buchanan, and Kristie Dotson. 2021.
\newblock Epistemic exclusion: Scholar (ly) devaluation that marginalizes
  faculty of color.
\newblock \emph{Journal of Diversity in Higher Education}, 14(4):493.

\bibitem[{Shaxobiddin(2024)}]{shaxobiddin2024discourse}
Abdullayev Shaxobiddin. 2024.
\newblock A discourse analysis of modal verbs in modern english: Patterns and
  functions.
\newblock \emph{Journal of new century innovations}, 50(2):145--147.

\bibitem[{Sinha et~al.(2020)Sinha, Parthasarathi, Wang, Lowe, Hamilton, and
  Pineau}]{sinha2020learning}
Koustuv Sinha, Prasanna Parthasarathi, Jasmine Wang, Ryan Lowe, William~L
  Hamilton, and Joelle Pineau. 2020.
\newblock Learning an unreferenced metric for online dialogue evaluation.
\newblock \emph{arXiv preprint arXiv:2005.00583}.

\bibitem[{Stolcke et~al.(2000)Stolcke, Ries, Coccaro, Shriberg, Bates,
  Jurafsky, Taylor, Martin, Ess-Dykema, and Meteer}]{stolcke2000dialogue}
Andreas Stolcke, Klaus Ries, Noah Coccaro, Elizabeth Shriberg, Rebecca Bates,
  Daniel Jurafsky, Paul Taylor, Rachel Martin, Carol~Van Ess-Dykema, and Marie
  Meteer. 2000.
\newblock Dialogue act modeling for automatic tagging and recognition of
  conversational speech.
\newblock \emph{Computational linguistics}, 26(3):339--373.

\bibitem[{Tao et~al.(2018)Tao, Mou, Zhao, and Yan}]{tao2018ruber}
Chongyang Tao, Lili Mou, Dongyan Zhao, and Rui Yan. 2018.
\newblock Ruber: An unsupervised method for automatic evaluation of open-domain
  dialog systems.
\newblock In \emph{Proceedings of the AAAI conference on artificial
  intelligence}, volume~32.

\bibitem[{Warren(2017)}]{warren2017cross}
Thomas Warren. 2017.
\newblock \emph{Cross-cultural Communication: Perspectives in theory and
  practice}.
\newblock Routledge.

\bibitem[{Whittaker et~al.(2021)Whittaker, Rogers, Petrovskaya, and
  Zhuang}]{10.1145/3424153}
Steve Whittaker, Yvonne Rogers, Elena Petrovskaya, and Hongbin Zhuang. 2021.
\newblock \href {https://doi.org/10.1145/3424153} {Designing personas for
  expressive robots: Personality in the new breed of moving, speaking, and
  colorful social home robots}.
\newblock \emph{J. Hum.-Robot Interact.}, 10(1).

\bibitem[{Wu and Roever(2021)}]{wu2021proficiency}
Jingxuan Wu and Carsten Roever. 2021.
\newblock Proficiency and preference organization in second language mandarin
  chinese refusals.
\newblock \emph{The Modern Language Journal}, 105(4):897--918.

\bibitem[{Yap and Sahoo(2024)}]{yap2024versatile}
Foong~Ha Yap and Anindita Sahoo. 2024.
\newblock Versatile copulas and their stance-marking uses in conversational
  odia, an indo-aryan language.
\newblock \emph{Lingua}, 297:103641.

\end{thebibliography}
\newpage
\onecolumn
\appendix

\section{Appendix} \label{sec:appendix}
\subsection{Software Availability} \label{subapp:SA}
To contribute to the research community and facilitate further development and collaboration, we have made the source codes of our innovative annotation tool publicly available. The tool, designed with a focus on enhancing the efficiency and accuracy of data annotation processes, has been developed through meticulous research and development efforts. It incorporates a range of features tailored to meet the needs of researchers and practitioners working in fields that require precise and reliable annotation of datasets.

\subsubsection*{Accessing the Source Code}
The source codes are hosted on GitHub, a platform widely recognized for its robust version control and collaborative features. Interested parties can access the repository at the following link: \url{https://anonymous.4open.science/r/AnnotationTool2023-CFE1/README.md}. This repository is intended for research usage, underlining our commitment to supporting academic and scientific endeavours.

\subsubsection*{Key Features and Capabilities}
Our annotation tool stands out for its user-friendly interface, which simplifies the annotation process and allows users to work more efficiently. Among its key features are:
\begin{itemize}
    \item \textbf{Customizable Annotation Labels:} Users can add their own set of labels to cater to the specific requirements of their projects.
    \item \textbf{Collaborative Annotation Support:} Facilitating teamwork, the tool allows multiple annotators to work on the same dataset simultaneously, ensuring consistency and reducing the time required for project completion.
    \item \textbf{Annotation History Tracking:} All the annotation history such as changes made can be tracked, and any further modifications can be done at any time for the user's convenience.
Export Functionality: Annotated data can be exported in several formats, accommodating further analysis or use in machine learning models.
\end{itemize}

\subsection{Pages View} \label{subapp:PV}
\begin{figure}[H]
\centerline{\includegraphics[width=1\linewidth]{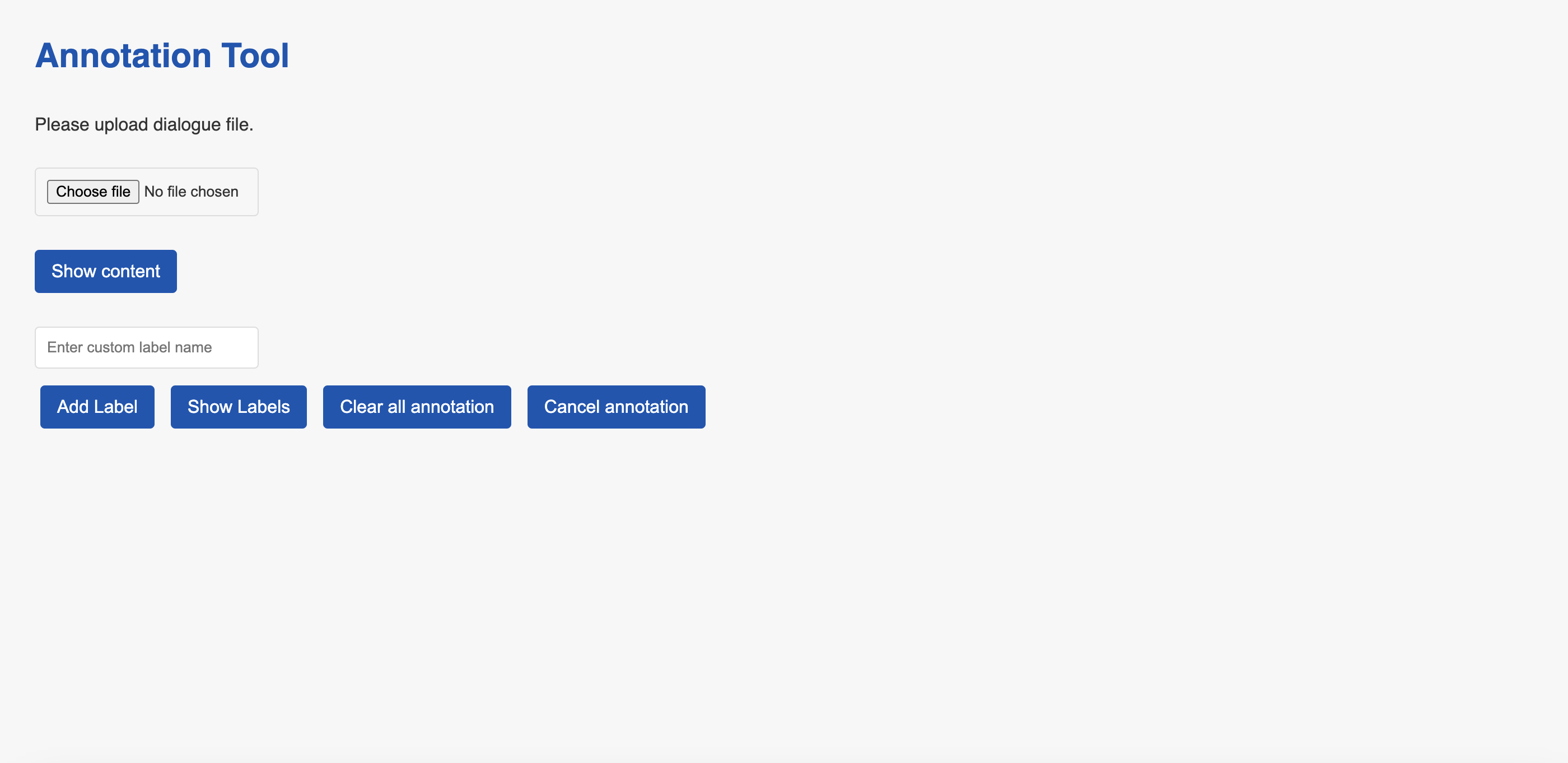}} 
\caption{Annotation tool Demo}
\label{fig:pageview1}
\end{figure}
\begin{figure}[H]
\centerline{\includegraphics[width=1\linewidth]{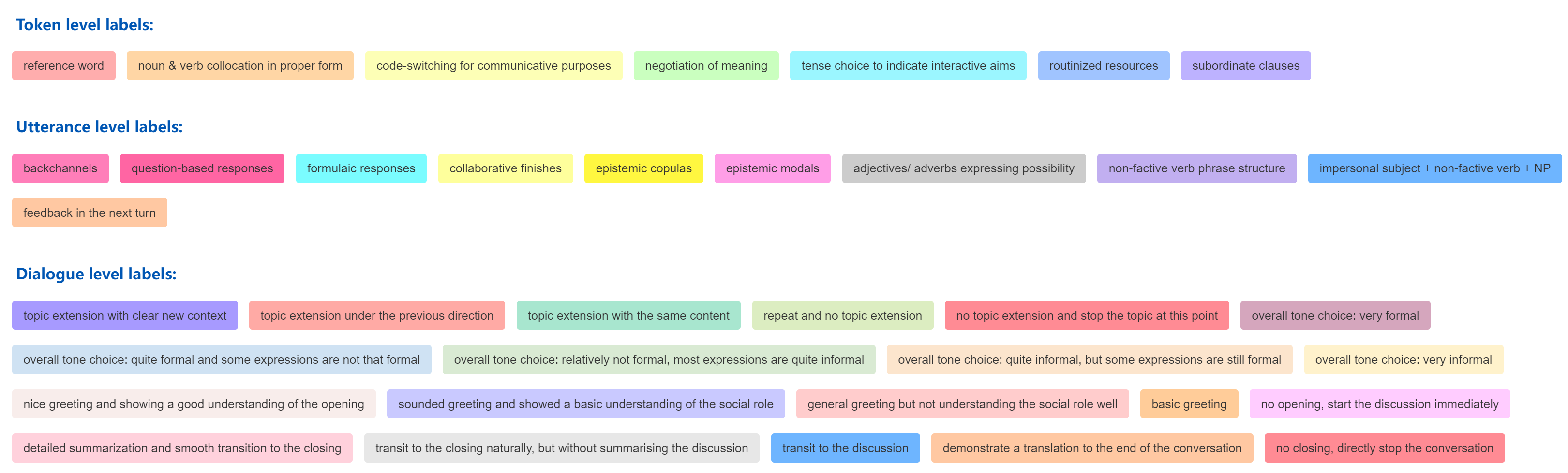}}
\caption{Hierarchical Label Assignment Demo}
\label{fig:pageview2}
\end{figure}





\includepdf[pages=1, scale=0.8, pagecommand={\subsection{Manual}\label{subapp:MN}}]{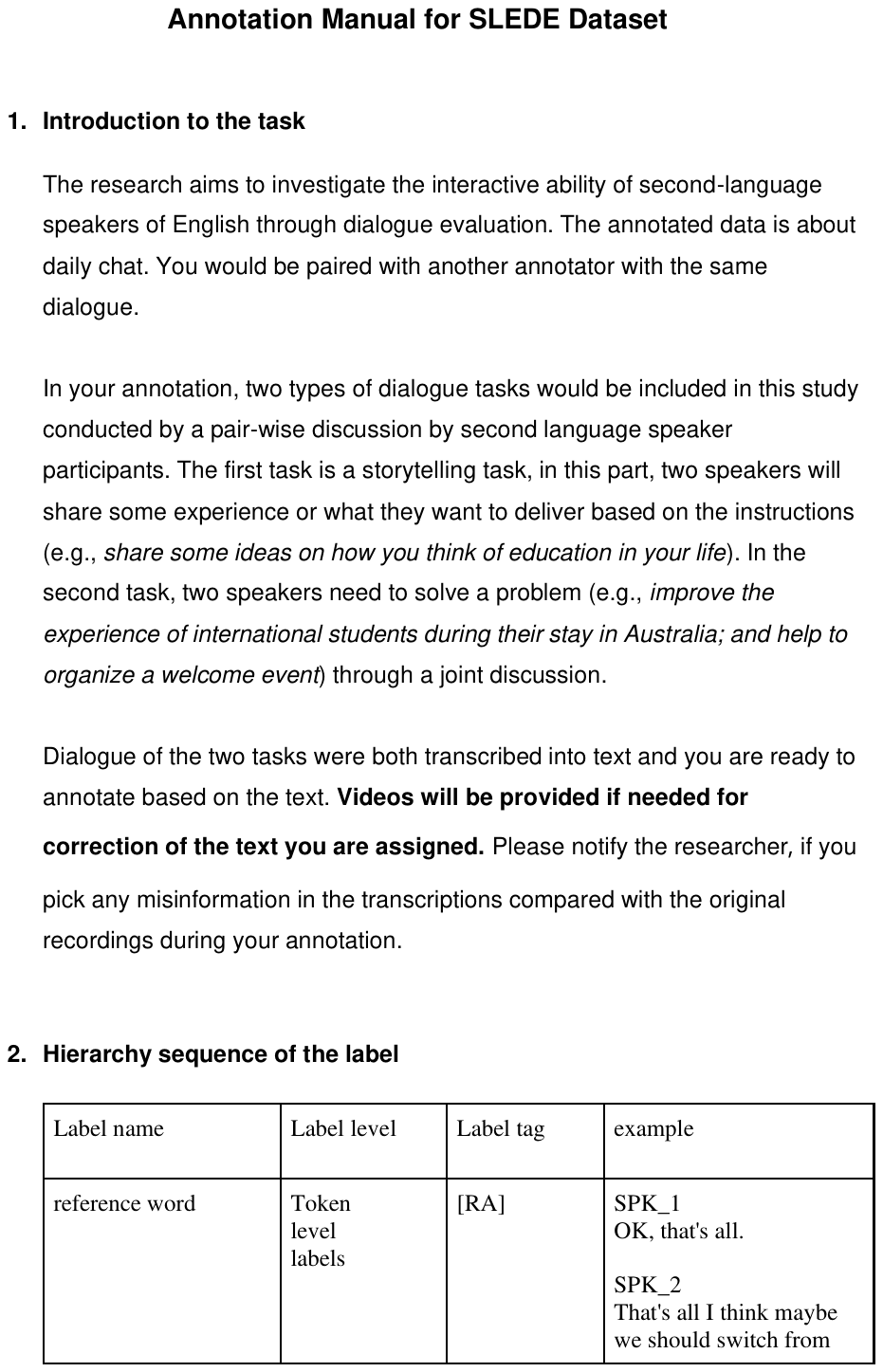}
\includepdf[pages=2-, scale=0.8, pagecommand={}]{appendix/Annotation_Munnal_2023.pdf}

\includepdf[pages=1, scale=0.8, pagecommand={\subsection{Questionnaire}\label{subapp:QA}}]
{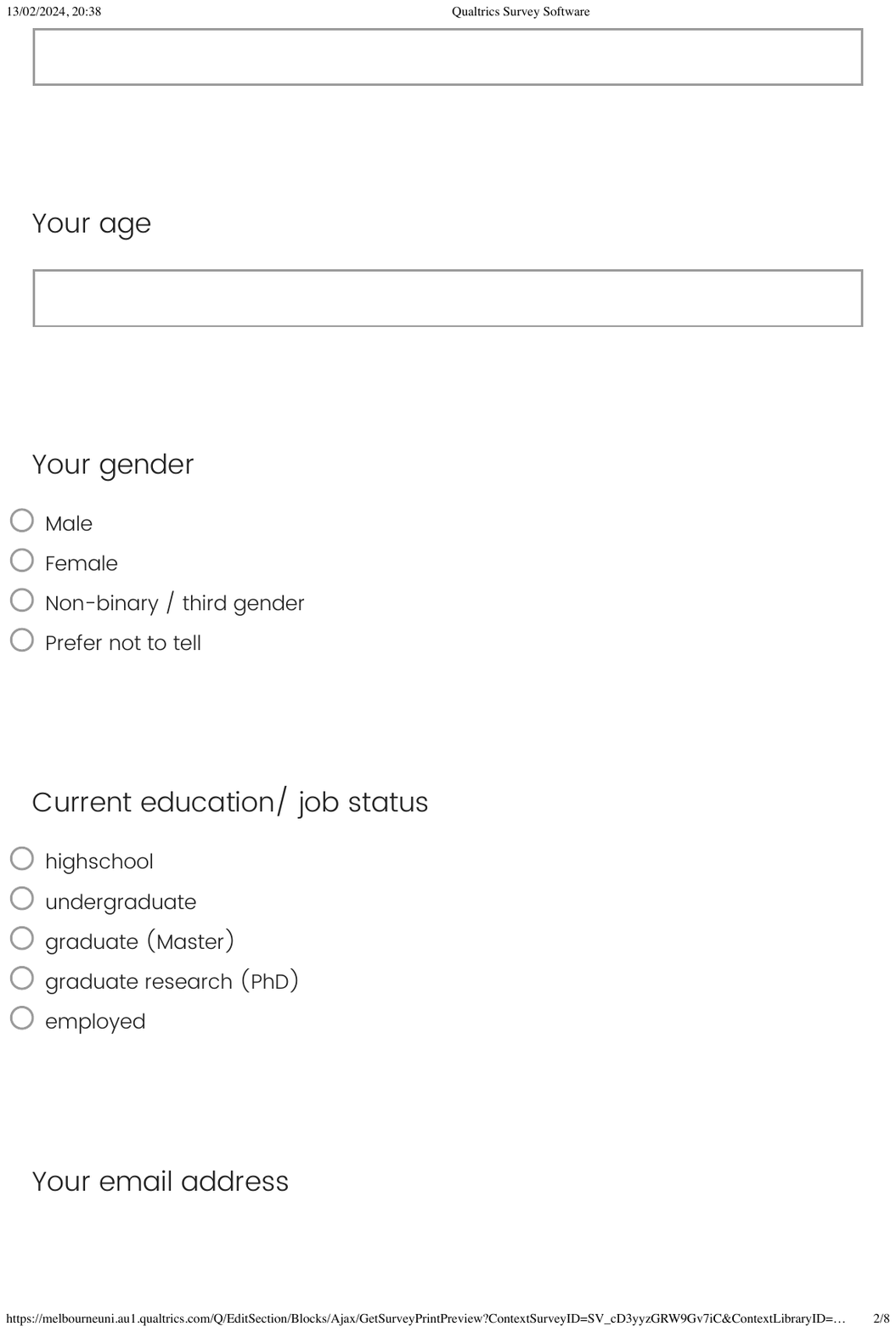}
\includepdf[pages=2-, scale=0.8, pagecommand={}]{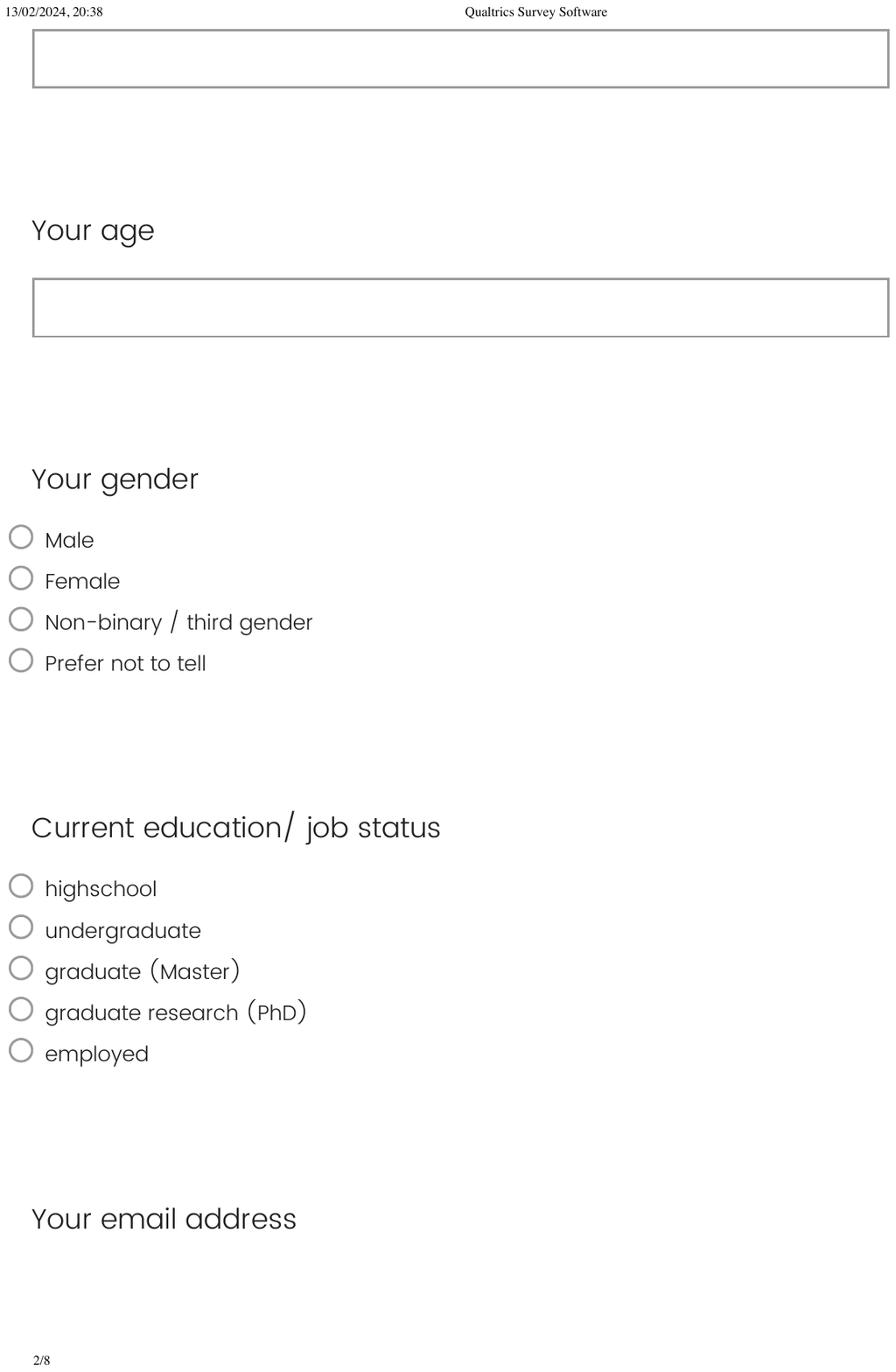}

\includepdf[pages=1, scale=0.8, pagecommand={\subsection{Speaking Instruments}\label{subapp:SI}}]
{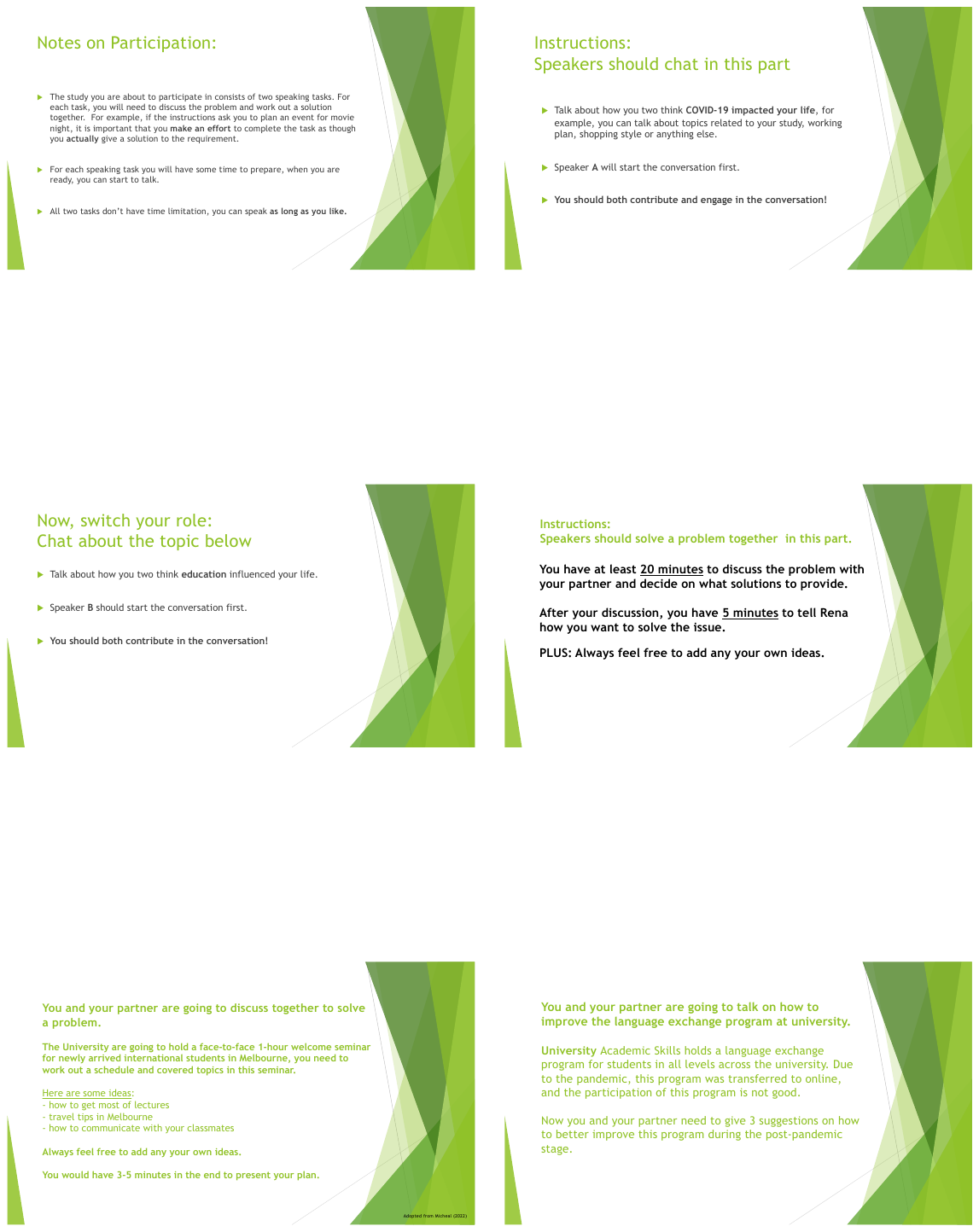}

\subsection{Experimental Result}\label{subapp:ER}
Since our research results include a large number of figures and extensive data, we have organized them into a thorough document available on our GitHub repository. This helps us keep the information accurate and detailed for in-depth examination. To view all the results, the readers can visit this link: \url{https://github.com/RenaGao/2024InteractiveMetrics/tree/main/2024ACLESLMainCodes_Results}. Storing the results in this way makes them easy to navigate and ensures the quality and precision of the research are maintained.


\end{document}